\documentclass[sigconf]{acmart}

\AtBeginDocument{%
  \providecommand\BibTeX{{%
    \normalfont B\kern-0.5em{\scshape i\kern-0.25em b}\kern-0.8em\TeX}}}

\copyrightyear{2021}
\acmYear{2021}
\setcopyright{acmcopyright}\acmConference[MM '21]{Proceedings of the 29th ACM
International Conference on Multimedia}{October 20--24, 2021}{Virtual Event, China}
\acmBooktitle{Proceedings of the 29th ACM International Conference on Multimedia
(MM '21), October 20--24, 2021, Virtual Event, China}
\acmPrice{15.00}
\acmDOI{10.1145/3474085.3475282}
\acmISBN{978-1-4503-8651-7/21/10}

\begin{document}
\fancyhead{}

\title{Neighbor-view Enhanced Model for Vision and Language Navigation}


\author{Dong An$^{1,2}$,\quad Yuankai Qi$^3$,\quad Yan Huang$^{1}$*,\quad Qi Wu$^{3}$, \quad Liang Wang$^{1,4,5}$, \quad Tieniu Tan$^{1,4}$}

\makeatletter
\def\authornotetext#1{
	\if@ACM@anonymous\else
	\g@addto@macro\@authornotes{
		\stepcounter{footnote}\footnotetext{#1}}
	\fi}
\makeatother
\authornotetext{Corresponding author.}

\affiliation{
	\institution{\textsuperscript{\rm 1}
	Center for Research on Intelligent Perception and Computing, Institution of Automation, Chinese Academy of Sciences
	\country{}}
}
\affiliation{
	\institution{\textsuperscript{\rm 2}School of Future Technology, University of Chinese Academy of Sciences, \textsuperscript{\rm 3}University of Adelaide
	\country{}}
}
\affiliation{
	\institution{\textsuperscript{\rm 4}
	Center for Excellence in Brain Science and Intelligence Technology (CEBSIT)
    \country{}}
}
\affiliation{
	\institution{\textsuperscript{\rm 5}
	 Chinese Academy of Sciences, Artificial Intelligence Research (CAS-AIR)
	\country{}}
}

\email{dong.an@cripac.ia.ac.cn,qykshr@gmail.com,{yhuang,wangliang,tnt}@nlpr.ia.ac.cn,qi.wu01@adelaide.edu.au}

\def\authors{Dong An, Yuankai Qi, Yan Huang, Qi Wu, Liang Wang, Tieniu Tan}

\renewcommand{\shortauthors}{An and Qi et al.}


\begin{abstract}
 	Vision and Language Navigation (VLN) requires an agent to navigate to a target location by following natural language instructions. Most of existing works represent a navigation candidate by the feature of the corresponding single view where the candidate lies in. However, an instruction may mention landmarks out of the single view as references, which might lead to failures of textual-visual matching of existing methods. In this work, we propose a multi-module \textit{Neighbor-View Enhanced Model} (NvEM) to adaptively incorporate visual contexts from neighbor views for better textual-visual matching. Specifically, our NvEM utilizes a subject module and a reference module to collect contexts from neighbor views. The subject module fuses neighbor views at a global level, and the reference module fuses neighbor objects at a local level. Subjects and references are adaptively determined via attention mechanisms. Our model also includes an action module to utilize the strong orientation guidance (e.g., ``turn left'') in instructions. Each module predicts navigation action separately and their weighted sum is used for predicting the final action. Extensive experimental results demonstrate the effectiveness of the proposed method on the R2R and R4R benchmarks against several state-of-the-art navigators, and NvEM even beats some pre-training ones. Our code is available at \url{https://github.com/MarSaKi/NvEM}.
\end{abstract}





\maketitle
\begin{figure*}[!tbp]
	\centering
	\includegraphics[width=0.9\textwidth]{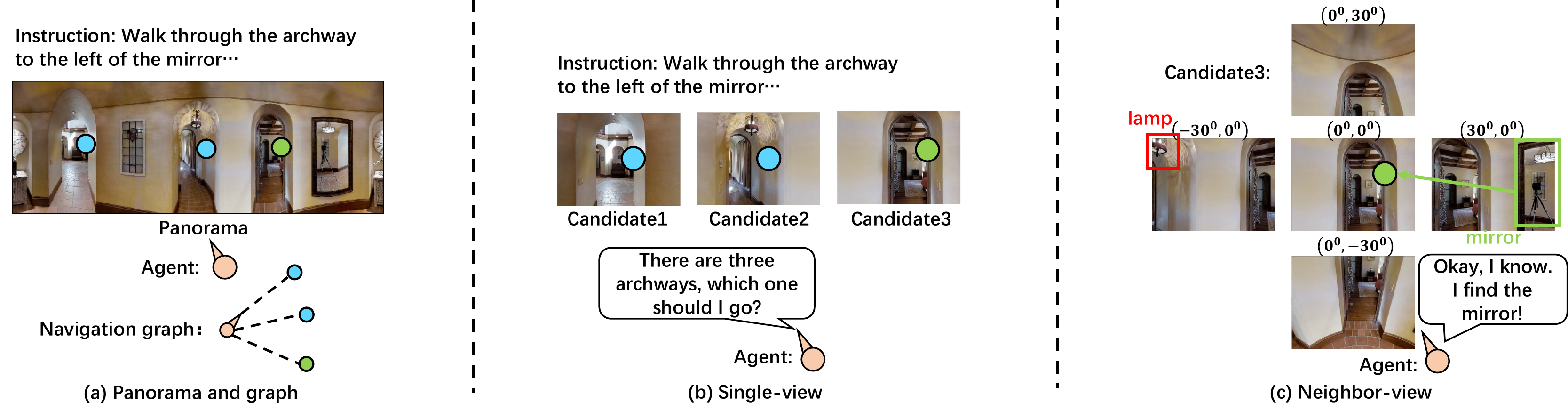}
	\vspace{-4mm}
	\caption{(a) Illustration of panorama view and navigation graph in VLN tasks. The yellow node is the current location. Candidate nodes are blue and green. Note that the green one is the ground-truth. (b) Single-view-based candidates used in previous works, which contain limited visual contexts and might be insufficient for action prediction. (c) Our neighbor-view scheme enriches the visual context of each candidate, leading to better textual-visual matching. Numbers above each view denote (heading, elevation).}
	\label{fig_ambiguity}
\end{figure*}
\begin{figure}[!tbp]
	\centering
	\includegraphics[width=0.95\linewidth]{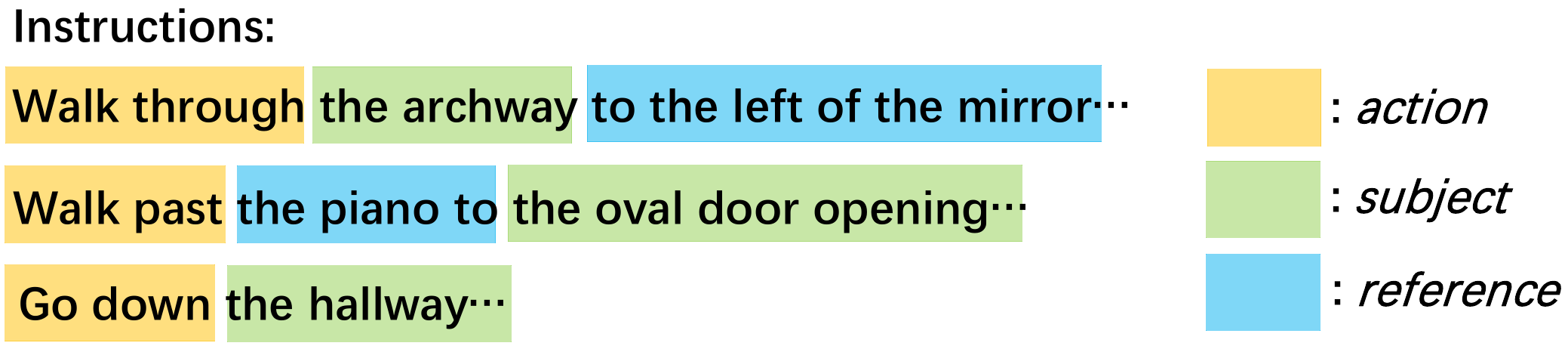}
	\vspace{-3mm}
	\caption{Examples of \textit{action}-, \textit{subject}- and \textit{reference}-related phrases in instructions. Some instructions do not contain \textit{references}.}
	\label{fig_gram_structure}
\end{figure}

\section{Introduction}\label{sec_introduction}
Vision and Language Navigation (VLN) has drawn increasing interest in recent years, partly because it represents a significant step towards enabling intelligent agents to interact with the realistic world. 
Running in a 3D simulator~\cite{anderson2018vln} rendered with real-world images~\cite{angle2017matterport}, the goal of VLN is to navigate to a target location by following a detailed natural language instruction, such as \textit{``Walk around the table and exit the room. Walk down the first set of stairs. Wait there.''}. There are two kinds of simulators, which render continuous navigation trajectories~\cite{jacob2020vlnce} and discrete trajectories~\cite{anderson2018vln} respectively. In this paper we focus on the discrete one, where the agent navigates on a discrete graph (see Figure~\ref{fig_ambiguity} (a)).
%

A variety of approaches have been proposed to address the VLN problem~\cite{wang2018leap,fried2018speaker,landi2019dynamic,tan2019envdrop,ma2019self-monitoring,wang2019rcm,li2019press,qi2020oaam,hao2020prevelant,arjun2020vln-bert,wang2020multitask,wang2020SERL,hong2020relgraph,deng2020EGP,hong2020vln-bert}. Most of them adopt panoramic action space~\cite{fried2018speaker}, where the agent select a navigable candidate from its observations to transport at each step.
However, the context of navigable candidates is rarely discussed in existing works, and the commonly used single-view candidates are of limited visual contexts which may hamper the matching between instructions and the visual representations of candidates.
Figure~\ref{fig_ambiguity} shows such an example, where there are three candidate \textit{``archways''}, each of which is represented by a single-view visual perception (Figure~\ref{fig_ambiguity} (b)). 
According to the instruction, only the archway \textit{``to the left of mirror''} leads to the correct navigation. 
However, most of existing agents may fail because they cannot find the referred \textit{``mirror''} in any single-view-based candidate. 
Thus, we propose to enhance the textual-visual matching by fusing visual information from candidates' neighbor views as shown in Figure~\ref{fig_ambiguity} (c), which is rarely explored before.
It is non-trivial to fuse neighbor views for visual contexts modeling, because many unmentioned visual clues exist which may interfere with the agent's decision (e.g., the lamp in Figure~\ref{fig_ambiguity} (c)). 
In addition, some instructions even do not involve visual clues in neighbor views, such as \textit{``go through the doorway''}. 
To handle this challenging problem, we propose to decompose an instruction into: \textit{action}-, \textit{subject}- and \textit{reference}-related phrases as shown in Figure~\ref{fig_gram_structure}.
Generally, the \textit{action} and \textit{subject} are necessary, and the optional \textit{reference} helps to distinguish the desired candidate from other similar ones. 
%

Based on the above mentioned three types of instruction phrases, we further design a multi-module \textit{Neighbor-view Enhanced Model} (NvEM) to adaptively fuse neighbor visual contexts in order to improve the textual-visual matching between instructions and candidates' visual perceptions. 
Specifically, our NvEM includes a subject module, a reference module and an action module,  where subjects and references are determined via attention mechanisms.
%
On one hand, the \textit{subject} module aggregates neighbor views at a global level based on spatial information. 
On the other hand, the \textit{reference} module aggregates related objects from neighbor views at a local level.
The action module makes use of the  orientation guidance (i.e., ``turn left'') in instructions.
Each module predicts navigation action separately and their weighted sum is used to predict the final action.
Note that the combination weights are trainable and predicted based on the decomposed \textit{subject-}, \textit{reference-} and \textit{action-}related phrases.

The contributions of this work are summarized as follows:

$\bullet$ To improve the textual-visual matching between instructions and navigable candidates, we propose to take into account the visual contexts from neighbor views for the first time.

$\bullet$ We propose a subject module and a reference module to adaptively fuse visual contexts from neighbor views at both global level and local level.

$\bullet$ Extensive experimental results demonstrate the effectiveness of the proposed method with comparisons against several existing state-of-the-art methods, and NvEM even beats some pre-training ones.

\vspace{-3mm}
\section{Related Work}

\noindent\textbf{Vision and Language Navigation.} Numerous approaches have been proposed to address the VLN problem.
Most of them are based on the CNN-LSTM architecture with attention mechanisms: at each time step, the agent first grounds surrounding observations to instructions, then chooses the most matched candidate according to the grounded instructions as the next location. 
Early work Speaker-Follower~\cite{fried2018speaker} develops a speaker model to synthesize new instructions for randomly sampled trajectories. 
Additionally, they design a panoramic action space for efficient navigation. 
Later on, EnvDrop~\cite{tan2019envdrop} increases the diversity of synthetic data by randomly removing objects to generate ``new environments''. 

On the other line, Self-monitoring~\cite{ma2019self-monitoring} and RCM~\cite{wang2019rcm} utilize the cross-modality co-attention mechanism to enhance the alignment of instructions and trajectories.
To learn generic linguistic and visual representations for VLN, AuxRN~\cite{zhu2020self-supervised} designs several auxiliary self-supervised losses. 
Very recently, large-scale pre-training models for VLN are widely explored~\cite{li2019press,hao2020prevelant,arjun2020vln-bert,hong2020vln-bert}, where they improve the agent's generation abilities dramatically by benefitting from priors of other datasets. 
Different types of visual clues correspond to different phrases in an instruction, OAAM~\cite{qi2020oaam} and RelGraph~\cite{hong2020relgraph} utilize decomposed phrases to guide more accurate action prediction.
OAAM~\cite{qi2020oaam} adopts action and object specialized clues to vote action at each time step, while RelGraph~\cite{hong2020relgraph} proposes a graph network to model the intra- and inter-relationships among the contextual and visual clues. 
The most relevant work to ours is RelGraph~\cite{hong2020relgraph}, where we both attempt to exploit view-level and object-level features. The key difference is: we focus on enhancing each candidate's representation with its multiple neighbor views (namely inter-view), while the representation in RelGraph is limited to a single view and thus it's intra-view. 

\noindent\textbf{Modular Attention Networks.} Modular networks are widely adopted in vision and language models. It attempts to decompose sentences into multiple phrases via attention mechanisms, as different phrases usually correspond to different visual clues.
MAttNet~\cite{yu2018mattnet} decomposes a long sentence into appearance, location and relationship three parts for referring expression. 
LCGN~\cite{hu2019lcgn} utilizes a multi-step textual attention mechanism to extract different object-related phrases, then modeling objects‘ contexts via the relations among the phrases.
LGI~\cite{mun2020lgi} utilizes a sequential query attention module to decompose the query into multiple semantic phrases, then using these phrases to interact with video clips for video grounding. 
To the best of our knowledge, OAAM~\cite{qi2020oaam} is the earliest attempt to decompose instructions in VLN. They decompose instructions into \textit{``action''} and \textit{``object ''} specialized phrases, and use these phrases to vote next action.
Our work has two key differences with OAAM: (I) Different modules: our modules are (subject, reference, action) \textit{v.s.} (object, action) of OAAM. (II) Our subject module and reference module fuse information from neighbor views while OAAM only use information within one single view.

\begin{figure*}[!htbp]
	\centering
	\includegraphics[width=\textwidth]{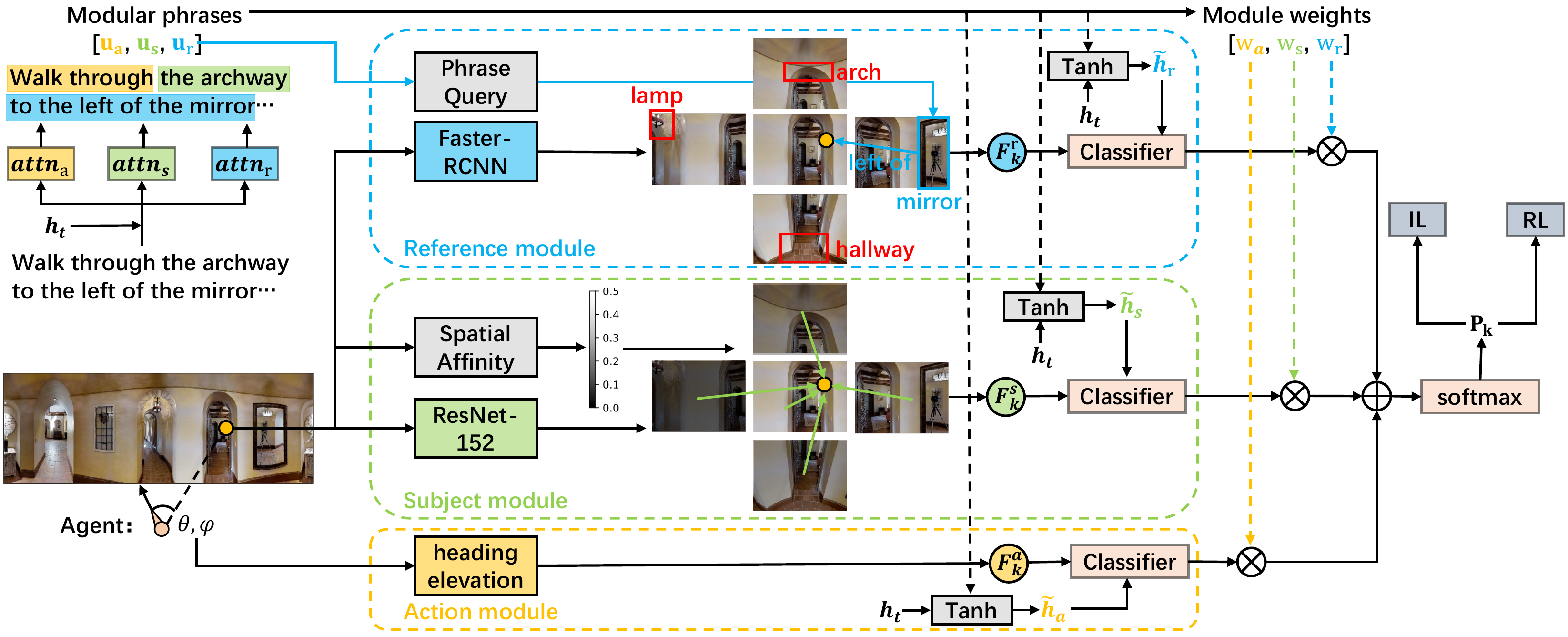}
	\vspace{-6mm}
	\caption{Main architecture of the proposed multi-module Neighbor-view Enhanced Model (NvEM).
	First, \textit{action}-, \textit{subject}- and \textit{reference}-related phrases are attended via an attention mechanism (Section~\ref{sec_extractor}). 
	Then, reference module and subject module predict navigation actions via aggregating visual contexts from candidates' neighbor views at local and global levels. The action module predicts navigation in terms of orientation information. (Section~\ref{sec_predictor}). 
	Lastly, the weighted sum of all three predictions predicts the final navigation decision. (Section~\ref{sec_integrator}).}
	\label{fig_framework}
\end{figure*}
\section{Preliminary}
In VLN~\cite{anderson2018vln}, given a natural language instruction with $L$ words, an agent navigates on a discrete graph to reach the described target by following the instruction. 
At each time step $t$, the agent observes a panorama which consists of 36 discrete views. 
Each view is represented by an image $\mathbf{v}_{t,i}$, with its orientation including heading $\mathbf{\theta}_{t,i}$ and elevation $\mathbf{\phi}_{t,i}$. 
Also, there are $K$ candidates at time step $t$, and each candidate is represented by the single view where the candidate lies in, with the candidate's relative orientation to the agent. Formally, for the $i$-th view:
\begin{equation}
	\begin{aligned}\label{form_angle}
		\mathbf{f}_{t,i}&=[\textrm{ResNet}(\mathbf{v}_{t,i}); E_{ori}(\mathbf{\theta}_{t,i}, \mathbf{\phi}_{t,i})]
	\end{aligned}
\end{equation}
where $\textrm{ResNet}(\cdot)$ represents ResNet~\cite{he2016resnet} pooling features, $E_{ori}(\cdot)$ is an embedding function for heading and elevation, which repeats $[\cos \mathbf{\theta}_{t,i},\sin \mathbf{\theta}_{t,i}$, $\cos\mathbf{\phi}_{t,i},\sin\mathbf{\phi}_{t,i}]$ 32 times following~\cite{tan2019envdrop}. The $k$-th candidate $\mathbf{g}_{t,k}$ is encoded in the same way.

Previous works show that data augmentation is able to significantly improve the generalization ability in unseen environments~\cite{fried2018speaker,tan2019envdrop}. 
We make use of this strategy by adopting EnvDrop~\cite{tan2019envdrop} as baseline, which first uses a bi-directional LSTM to encode the instruction, and the encoded instruction is represented as $\mathbf{I}=\{\mathbf{u_l}\}_{l=1}^L$.
Then the agent's previous context-aware state $\tilde{\mathbf{h}}_{t-1}$ is used to attend on all views to get scene feature: $\tilde{\mathbf{f}}_t=\textrm{SoftAttn}(\tilde{\mathbf{h}}_{t-1},\{\mathbf{f}_{t,i}\}_{i=1}^{36})$.
The concatenation of $\tilde{\mathbf{f}}_t$ and the previous action embedding $\mathbf{a}_{t-1}$ is fed into the decoder LSTM to update the agent's state: $\mathbf{h}_t=\textrm{LSTM}([\tilde{\mathbf{f}}_t;\mathbf{a}_{t-1}],\tilde{\mathbf{h}}_{t-1})$. 
Note that the context-aware agent state is updated via the attentive instruction feature $\mathbf{u}_t$:
\begin{equation}
	\begin{aligned}\label{form_ht}
		\tilde{\mathbf{h}}_t=\textrm{Tanh}(\mathbf{W}_H[\mathbf{u}_t;\mathbf{h}_t]), \ \
		\mathbf{u}_t=\textrm{SoftAttn}(\tilde{\mathbf{h}}_{t-1},\{\mathbf{u}_{l}\}_{l=1}^{L})
	\end{aligned}
\end{equation}
where $\mathbf{W}_H$ is a trainable linear projection and $\textrm{SoftAttn}(\cdot)$ mentioned above denotes soft-dot attention.
Finally, EnvDrop predicts navigation action by selecting the candidate with the highest probability ($\mathbf{W}_G$ is a trainable linear projection):
\begin{align}\label{form_select_action}
	\mathbf{a}_t^*=\mathop{\arg\max}_k P(\mathbf{g}_{t,k}), \ \ P(\mathbf{g}_{t,k})=\textrm{softmax}_k(\mathbf{g}_{t,k}^\top \mathbf{W}_G \tilde{\mathbf{h}}_t)
\end{align}

\section{Methodology}
In this section, we first briefly describe the pipeline of the proposed model in training phase, then we detail the proposed Neighbor-view Enhanced Model (NvEM). 
Note that in this section, we omit the time step $t$ to avoid notational clutter in the exposition.

\subsection{Overview}
Figure~\ref{fig_framework} illustrates the main pipeline of our NvEM. First, \textit{action}-, \textit{subject}- and \textit{reference}-related phrases are attended by three independent attention schemes. 
Then, reference module and subject module predict navigation actions via aggregating visual contexts from candidates' neighbor views. The action module predicts navigation in terms of orientation information.
Lastly, the final navigation action is determined by combining these three predictions together with weights generated from phrases' embeddings.

\subsection{Phrase Extractor}\label{sec_extractor}
Considering an instruction, such as \textit{``walk through the archway to the left of the mirror...''}, there are three types of phrases which the agent needs to identify: the \textit{action} describes the orientation of target candidate (e.g., \textit{``walk through''}), the \textit{subject} describes the main visual entity of the correct navigation (e.g., \textit{``the archway''}) and the \textit{reference} which is referenced by the subject (e.g., \textit{``to the left of the mirror''}). 
Thus, NvEM first performs three soft-attentions independently on the instruction, conditioned on the current agent state $\mathbf{h}_t$, to attend on these three types of phrases:
\begin{equation}
	\begin{aligned}
		&\tilde{\mathbf{h}}_a=\textrm{Tanh}(\mathbf{W}_{Ha}[\mathbf{u}_a;\mathbf{h}_t]),\ \
		\mathbf{u}_a=\textrm{SoftAttn}_a(\tilde{\mathbf{h}}_{t-1},\{\mathbf{u}_{l}\}_{l=1}^{L}) \\
		&\tilde{\mathbf{h}}_s=\textrm{Tanh}(\mathbf{W}_{Hs}[\mathbf{u}_s;\mathbf{h}_t]),\ \
		\mathbf{u}_s=\textrm{SoftAttn}_s(\tilde{\mathbf{h}}_{t-1},\{\mathbf{u}_{l}\}_{l=1}^{L}) \\
		&\tilde{\mathbf{h}}_r=\textrm{Tanh}(\mathbf{W}_{Hr}[\mathbf{u}_r;\mathbf{h}_t]),\ \
		\mathbf{u}_r=\textrm{SoftAttn}_r(\tilde{\mathbf{h}}_{t-1},\{\mathbf{u}_{l}\}_{l=1}^{L})
	\end{aligned}
\end{equation}
where the subscripts denote the corresponding types of phrases, $\textrm{SoftAttn}_*(\cdot)$ is the same with Eq~\eqref{form_ht}. $\mathbf{u}_*$ and $\mathbf{\tilde{h}}_*$ denote features of corresponding phrases and context-aware agent states, and they are updated by different linear projections $\mathbf{W}_{H*}$. The global context-aware agent state $\mathbf{\tilde{h}}_t$ in Eq~\eqref{form_ht} is now calculated by averaging the three specialized context-aware states $\mathbf{\tilde{h}}_a$, $\mathbf{\tilde{h}}_s$ and $\mathbf{\tilde{h}}_r$.

\subsection{Neighbor-view Enhanced Navigator}
\label{sec_predictor}
Corresponding to the attended three types of phrases in instructions, our neighbor-view enhanced navigator contains three modules: a reference module, a subject module and an action module. The reference module and subject module predict navigation actions via aggregating visual contexts from neighbor views at local and global levels, respectively. The action module predicts navigation actions according to orientation information. We provide the details below. 

\vspace{1mm}
\noindent\textbf{Reference Module}. \textit{Reference} usually exists as a landmark surrounding the \textit{subject} to clarify similar navigation candidates. 
In the example \textit{``walk through the archway to the left of the mirror...''}, the \textit{reference} is \textit{``the mirror''} and it is referred with a spatial relationship to the \textit{subject} (e.g., \textit{``to the left of''}). 
This motivates us to enhance the representation of a navigation candidate by using features of local objects and their spatial relations.

Formally, for the $k$-th candidate $\mathbf{g}_{k}$, given its orientation $\mathbf{d}_k$, the objects' features $\{\mathbf{o}_{k,i,j}\}_{j=1}^J$ extracted by~\cite{ren2015faster-rcnn} and their orientations $\{\mathbf{d}_{k,i,j}\}_{j=1}^J$ in the candidate's neighbor views (assume there are $J$ objects in each neighbor and $i$ denotes the $i$-th neighbor). 
We then calculate the spatial relations of the neighbor objects to the $k$-th candidate:
\begin{align}\label{form_rel_anlge}
    \mathbf{e}_{k,i,j}=E_{ori}(\mathbf{d}_{k,i,j}-\mathbf{d}_k)
\end{align}
where $E_{ori}(\cdot)$ is the same as that in Eq~\eqref{form_angle}. Then each neighbor object is represented as the concatenation of its object feature and relative spatial embedding:
\begin{align}
    \mathbf{\bar{o}}_{k,i,j}=\mathbf{\Gamma}_o([\mathbf{o}_{k,i,j};\mathbf{e}_{k,i,j}])
\end{align}
where $[:]$ denotes concatenation, $\mathbf{\Gamma}_o(\cdot)$ projects objects' features into a $\mathbf{D}_k$ dimensional space and all $\mathbf{\Gamma}(\cdot)_*$ in this section represent trainable non-linear projections, with $\textrm{Tanh}(\cdot)$ as the activation function. We use the reference-related phrases $\mathbf{u}_r$ to highlight relevant objects in neighbor views:
\begin{equation}
	\begin{aligned}
		\mathbf{A}_{k,i,j}^r=\textrm{softmax}_{i,j}&(\frac{(\mathbf{W}_q^r\cdot \mathbf{u}_r)^\top(\mathbf{W}_k^r\cdot \mathbf{\bar{o}}_{k,i,j})}{\sqrt{\mathbf{D}_k}})\\
		\mathbf{F}_k^r=&\sum_{i,j}\mathbf{A}_{k,i,j}^r\cdot \tilde{\mathbf{o}}_{k,i,j}
	\end{aligned}
\end{equation}
where $\mathbf{W}_q^r$ and $\mathbf{W}_k^r$ are trainable linear projections. The \textit{reference} module predicts the confidence of the candidate $\mathbf{g}_k$ being the next navigation action using \textit{reference}-related state
$\mathbf{\tilde{h}}_r$ and the neighbor reference enhanced candidate representation $\mathbf{F}_k^r$:
\begin{equation}
    \mathbf{\tau}_{r,k} = \mathbf{\tilde{h}}_r^\top \mathbf{\widetilde{W}}_r \mathbf{F}_k^r
\end{equation}
where $\mathbf{\widetilde{W}}_r$ is a trainable linear projection.

\vspace{1mm}
\noindent\textbf{Subject Module}. Instruction \textit{subject} describes the main visual entity of the correct navigation, such as \textit{``the archway''} in \textit{``walk through the archway to the left of the mirror...''}. 
However, sometimes there are multiple candidates containing the different instances of the \textit{subject}.
To alleviate the ambiguity in visual side, we propose to enhance the visual representation of subject by incorporating contexts from neighbor views.
Specifically, we aggregate neighbor views at a global level, with the help of the spatial affinities neighbor views to the candidate.

Formally, for the $k$-th candidate $\mathbf{g}_k$, given its orientation $\mathbf{d}_k$, its neighbor views' orientations $\{\mathbf{d}_{k,i}\}_{i=1}^I$ and  ResNet features $\{\mathbf{v}_{k,i}\}_{i=1}^I$ (assume it has $I$ neighbor views), we first embed all neighbor views using a trainable non-linear projection:
\begin{align}
	\mathbf{\tilde{v}}_{k,i} = \mathbf{\Gamma}_s({\mathbf{v}}_{k,i})
\end{align}
Then we compute the spatial affinities among the candidate and its neighbor views based on their orientations in a query-key manner~\cite{ashish2017transformer}. Then the enhanced subject visual representation $\mathbf{F}_k^s$ is obtained by adaptively aggregating neighbor views' embeddings:
\begin{equation}
	\begin{aligned}\label{form_query}
		\mathbf{A}_{k,i}^s=\textrm{softmax}_i&(\frac{(\mathbf{W}_q^s\cdot \mathbf{d}_{k})^\top(\mathbf{W}_k^s\cdot \mathbf{d}_{k,i})}{\sqrt{\mathbf{D}_k}}) \\
		\mathbf{F}_k^s&=\sum_i \mathbf{A}_{k,i}^s\cdot \mathbf{\tilde{v}}_{k,i}
	\end{aligned}
\end{equation}
where $\mathbf{W}_q^s$ and $\mathbf{W}_k^s$ are trainable linear projections. 
Similar to \textit{reference} module, the \textit{subject} module predicts the confidence of candidate $\mathbf{g}_k$ being the next navigation action via: 
\begin{equation}
    \mathbf{\tau}_{s,k} = \mathbf{\tilde{h}}_s^\top \mathbf{\widetilde{W}}_s \mathbf{F}_k^s
\end{equation}
where $\mathbf{\widetilde{W}}_s$ is a trainable linear projection.

\vspace{1mm}
\noindent\textbf{Action Module}. \textit{Action} related phrases serve as strong guidance in navigation~\cite{hu2019looking,qi2020oaam}, such as \textit{``go forward''}, \textit{``turn left''} and \textit{``go down''}. Inspired by~\cite{qi2020oaam}, for the $k$-th candidate $\mathbf{g}_k$, the action module predicts the confidence of candidate $\mathbf{g}_k$ being the next navigation action, using the candidate's orientation $\mathbf{d}_k$ and action-related state $\tilde{h}_a$:
\begin{equation}
\begin{aligned}
\mathbf{F}_k^a =& \mathbf{\Gamma}_a(E_{ori}(\mathbf{d}_k))\\ 
	\mathbf{\tau}_{a,k} =&\mathbf{\tilde{h}}_a^\top \mathbf{\widetilde{W}}_a \mathbf{F}_k^a
\end{aligned}
\end{equation}
where $\mathbf{\widetilde{W}}_a$ is a trainable linear projection.

\subsection{Adaptive Action Integrator}\label{sec_integrator}
The \textit{action}, \textit{subject} and \textit{reference} modules usually contribute at different degrees to the final decision.
%
Thus, we propose to adaptively integrate predictions from these three modules. We first calculate the combination weights conditioned on \textit{action}-, \textit{subject}- and \textit{reference}-specialized phrases $\mathbf{u}_a$, $\mathbf{u}_s$ and $\mathbf{u}_r$. 
Take the \textit{action} weight $\mathbf{w}_a$ as an example:
\begin{align}
	\mathbf{w}_a = \frac{exp(\mathbf{W}_a\cdot \mathbf{u}_a)}{exp(\mathbf{W}_a\cdot \mathbf{u}_a)+exp(\mathbf{W}_s\cdot \mathbf{u}_s)+exp(\mathbf{W}_r\cdot \mathbf{u}_r)}
\end{align} 
where the $\mathbf{W}_a$, $\mathbf{W}_s$ and $\mathbf{W}_r$ are trainable linear projections. Then the final action probability of the $k$-th candidate is calculated by weighted summation of above confidences:
\begin{align}\label{form_fusion_weight}
	\mathbf{P}_k = \textrm{softmax}_k(\mathbf{w}_a\cdot \mathbf{\tau}_{a,k} + \mathbf{w}_s\cdot \mathbf{\tau}_{s,k} + \mathbf{w}_r\cdot \mathbf{\tau}_{r,k})
\end{align}
In the inference phase, the agent selects the candidate with the maximum probability as shown in Eq~\eqref{form_select_action} at each step.

\subsection{Training}
We apply the Imitation Learning (IL) + Reinforcement Learning (RL) objectives to train our model following~\cite{tan2019envdrop}. 
In imitation learning, the agent takes the teacher action $\mathbf{a}_t^*$ at each time step to learn to follow the ground-truth trajectory. 
In reinforcement learning, the agent samples an action $\mathbf{a}_t^S$ via the probability $\mathbf{P}_t$ and learns from the rewards. Formally:
\begin{align}
	\mathbf{L}=\mathbf{\lambda} \sum_{t=1}^{\mathbf{T}^*} -\mathbf{a}_t^*\log(\mathbf{P}_t) + \sum_{t=1}^{\mathbf{T}^S} -\mathbf{a}_t^S \log(\mathbf{P}_t) \mathbf{A}_t
\end{align}
where $\mathbf{\lambda}$ is a coefficient for weighting the IL loss, $\mathbf{T}^*$ and $\mathbf{T}^S$ are the total numbers of steps the agent takes in IL and RL respectively.
$\mathbf{A}_t$ is the advantage in A2C algorithm~\cite{volo2016a3c}. 
We apply the summation of two types of rewards in RL objective, goal reward and fidelity reward following~\cite{jain2019r4r}. 
%

\begin{table*}[htbp]
	\caption{Comparison of single-run performance with the state-of-the-art methods on R2R. $\dag$ denotes works that apply pre-trained textual or visual encoders.}
	\label{table_r2r}
	\vspace{-3mm}
	\centering
	\resizebox{0.9\textwidth}{!}{\begin{tabular}{lrrrrrrrrrrrr}
			\toprule 
			\multicolumn{1}{c}{} & \multicolumn{4}{c}{Val Seen} & \multicolumn{4}{c}{Val Unseen} & \multicolumn{4}{c}{Test Unseen} \\
			\cmidrule(r){2-5} \cmidrule(r){6-9} \cmidrule(r){10-13}
			\multicolumn{1}{c}{Agent} & \multicolumn{1}{c}{TL} & \multicolumn{1}{c}{NE$\downarrow$} & \multicolumn{1}{c}{SR$\uparrow$} & \multicolumn{1}{c}{SPL$\uparrow$} & \multicolumn{1}{c}{TL} & \multicolumn{1}{c}{NE$\downarrow$} & \multicolumn{1}{c}{SR$\uparrow$} & \multicolumn{1}{c}{SPL$\uparrow$} & \multicolumn{1}{c}{TL} & \multicolumn{1}{c}{NE$\downarrow$} & \multicolumn{1}{c}{SR$\uparrow$} & \multicolumn{1}{c}{SPL$\uparrow$} \\
			\midrule
			Random                & 9.58  & 9.45 & 0.16 & -    & 9.77  & 9.23 & 0.16 & -    & 9.89  & 9.79 & 0.13 & 0.12 \\
			Human                 & -     & -    & -    & -    & -     & -    & -    & -    & 11.85 & 1.61 & 0.86 & 0.76 \\
			\midrule
			PRESS~\cite{li2019press} $\dag$
			& 10.57 & 4.39 & 0.58 & 0.55 & 10.36 & 5.28 & 0.49 & 0.45 & 10.77 & 5.49 & 0.49 & 0.45 \\
			PREVALENT~\cite{hao2020prevelant} $\dag$
			& 10.32 & 3.67 & 0.69 & 0.65 & 10.19 & 4.71 & 0.58 & 0.53 & 10.51 & 5.30 & 0.54 & 0.51 \\
			VLNBert (init. OSCAR)~\cite{hong2020vln-bert}$\dag$	& 10.79 & 3.11 & 0.71 & 0.67 & 11.86 & 4.29 & 0.59 & 0.53 & 12.34 & 4.59 & 0.57 & 0.53 \\
			VLNBert (init. PREVALENT)~\cite{hong2020vln-bert}$\dag$ &  11.13 & 2.90 & 0.72 & 0.68 & 12.01 & 3.93 & 0.63 & 0.57 & 12.35 & 4.09 & 0.63 & 0.57 \\
			\midrule
			Seq2Seq~\cite{anderson2018vln}
			& 11.33 & 6.01 & 0.39 & -    & 8.39  & 7.81 & 0.22 & -    & 8.13  & 7.85 & 0.20 & 0.18 \\
			Speaker-Follower~\cite{fried2018speaker}
			& -     & 3.36 & 0.66 & -    & -     & 6.62 & 0.35 & -    & 14.82 & 6.62 & 0.35 & 0.28 \\
			SM~\cite{ma2019self-monitoring}
			& -     & \textbf{3.22} & 0.67 & 0.58 & -     & 5.52 & 0.45 & 0.32 & 18.04 & 5.67 & 0.48 & 0.35 \\
			RCM+SIL~\cite{wang2019rcm}
			& 10.65 & 3.53 & 0.67 & -    & 11.46 & 6.09 & 0.43 & -    & 11.97 & 6.12 & 0.43 & 0.38 \\
			Regretful~\cite{ma2019regretful}
			& -     & 3.23 & 0.69 & 0.63 & -     & 5.32 & 0.50 & 0.41 & 13.69 & 5.69 & 0.48 & 0.40 \\
			VLNBert (no init.)~\cite{hong2020vln-bert} & 9.78 & 3.92 & 0.62 & 0.59 & 10.31 & 5.10 & 0.50 & 0.46 & 11.15 & 5.45 & 0.51 & 0.47 \\
			EnvDrop~\cite{tan2019envdrop}
			& 11.00 & 3.99 & 0.62 & 0.59 & 10.70 & 5.22 & 0.52 & 0.48 & 11.66 & 5.23 & 0.51 & 0.47 \\
			AuxRN~\cite{zhu2020self-supervised}
			& -     & 3.33 & \textbf{0.70} & \textbf{0.67} & -     & 5.28 & 0.55 & 0.50 & -     & 5.15 & 0.55 & 0.51 \\
			RelGraph~\cite{hong2020relgraph}                  & 10.13 & 3.47 & 0.67 & 0.65 & 9.99  & 4.73 & 0.57 & 0.53 & 10.29 & 4.75 & 0.55 & 0.52 \\
			NvEM (ours) & 11.09 & 3.44 & 0.69 & 0.65 & 11.83 & \textbf{4.27} & \textbf{0.60} & \textbf{0.55} & 12.98 & \textbf{4.37} & \textbf{0.58} & \textbf{0.54} \\
			\bottomrule
	\end{tabular}}
\end{table*}

\begin{table*}[t]
	\caption{Comparison of single-run performance with the state-of-the-art methods on R4R. \textit{goal} and \textit{fidelity} indicate goal and fidelity reward in reinforcement learning. $\ddag$ denotes our reimplemented R4R results.}
	\label{table_r4r}
	\vspace{-3mm}
	\centering
	\resizebox{0.9\textwidth}{!}{\begin{tabular}{lrrrrrrrrrrrr}
			\toprule 
			\multicolumn{1}{c}{} & \multicolumn{6}{c}{Val Seen} & \multicolumn{6}{c}{Val Unseen} \\
			\cmidrule(r){2-7} \cmidrule(r){8-13}
			\multicolumn{1}{c}{Agent} & \multicolumn{1}{c}{NE$\downarrow$} & \multicolumn{1}{c}{SR$\uparrow$} & \multicolumn{1}{c}{SPL$\uparrow$} & \multicolumn{1}{c}{CLS$\uparrow$} & \multicolumn{1}{c}{nDTW$\uparrow$} & \multicolumn{1}{c}{sDTW$\uparrow$} & \multicolumn{1}{c}{NE$\downarrow$} & \multicolumn{1}{c}{SR$\uparrow$} & \multicolumn{1}{c}{SPL$\uparrow$} & \multicolumn{1}{c}{CLS$\uparrow$} & \multicolumn{1}{c}{nDTW$\uparrow$} & \multicolumn{1}{c}{sDTW$\uparrow$}\\
			\midrule
			EnvDrop~\cite{tan2019envdrop} & - & 0.52 & 0.41 & 0.53 & - & 0.27 & - & 0.29 & 0.18 & 0.34 & - & 0.09 \\
			RCM-a (goal)~\cite{jain2019r4r}    
			& \textbf{5.11} & \textbf{0.56} & 0.32 & 0.40 & -    & -    & 8.45 & 0.29 & 0.10 & 0.20 & -    & -    \\
			RCM-a (fidelity)~\cite{jain2019r4r} 
			& 5.37 & 0.53 & 0.31 & \textbf{0.55} & -    & -    & 8.08 & 0.26 & 0.08 & 0.35 & -    & -    \\
			RCM-b (goal)~\cite{grabriel2019general}    
			& -    & -    & -    & -    & -    & -    & -    & 0.29 & 0.15 & 0.33 & 0.27 & 0.11 \\
			RCM-b (fidelity)~\cite{grabriel2019general}    
			& -    & -    & -    & -    & -    &  -   & -    & 0.29 & 0.21 & 0.35 & 0.30 & 0.13 \\
			OAAM~\cite{qi2020oaam} & - & \textbf{0.56} & 0.49 & 0.54 & - & 0.32 & - & 0.31 & 0.23 & 0.40 & - & 0.11 \\
			RelGraph~\cite{hong2020relgraph}$\ddag$  & 5.14 & 0.55 & \textbf{0.50} & 0.51 & \textbf{0.48} & \textbf{0.35} & 7.55 & 0.35 & 0.25 & 0.37 & 0.32 & 0.18 \\
			\midrule
			NvEM (ours) & 5.38 & 0.54 & 0.47 & 0.51 & \textbf{0.48} & \textbf{0.35} & \textbf{6.85} & \textbf{0.38} & \textbf{0.28} & \textbf{0.41} & \textbf{0.36} & \textbf{0.20} \\ 
			\bottomrule
	\end{tabular}}
\end{table*}

\section{Experiments}
In this section, we first describe the commonly used VLN datasets and the evaluation metrics. 
Then we present implementation details of NvEM.
Finally, we compare against several state-of-the-art methods and provide ablation experiments. Qualitative visualizations are also presented.

\subsection{Datasets and Evaluation Metrics}
\noindent\textbf{R2R benchmark}. The Room-to-Room (R2R) dataset~\cite{anderson2018vln} consists of 10,567 panoramic view nodes in 90 real-world environments as well as 7,189 trajectories described by three natural language instructions. 
The dataset is split into train, validation seen, validation unseen and test unseen sets. We follow the standard metrics employed by previous works to evaluate the performance of our agent. These metrics include: the Trajectory Length (TL) which measures the average length of the agent's navigation path, the Navigation Error (NE) which is the average distance between the agent's final location and the target, the Success Rate (SR) which measures the ratio of trajectories where the agent stops at 3 meters within the target, and the Success Rate weighted by Path Length (SPL) which considers both path length and success rate. Note that the \textbf{SR} and \textbf{SPL} in unseen environments are main metrics for R2R.

\noindent\textbf{R4R benchmark}. The Room-for-Room (R4R) dataset~\cite{jain2019r4r} is an extended version of R2R, which has longer instructions and trajectories. 
The dataset is split into train, validation seen, validation unseen sets. 
Besides the main metrics in R2R, R4R includes additional metrics: the Coverage Weighted by Length Score (CLS)~\cite{jain2019r4r}, the Normalized Dynamic Time Wrapping (nDTW)~\cite{grabriel2019general} and the nDTW weighted by Success Rate (sDTW)~\cite{grabriel2019general}. In R4R, {SR} and {SPL} measure the accuracy of navigation, while {CLS}, {nDTW} and {sDTW} measure the fidelity of predicted paths and ground-truth paths.

\subsection{Implementation Details}
We use ResNet-152~\cite{he2016resnet} pre-trained on Places365~\cite{zhou2018places} to extract view features. 
We apply Faster-RCNN~\cite{ren2015faster-rcnn} pre-trained on the Visual Genome Dataset~\cite{ranjay2017genome} to obtain object labels in \textit{reference} module, then encoded by Glove~\cite{jeffrey2014glove}. 
To simplify the object vocabulary, we retain the top 100 most frequent classes mentioned in R2R training data following~\cite{hong2020relgraph}. 
Our method exploits neighbor views to represent a navigation candidate, thus the number of neighbors and objects are crucial for the final performance. 
Our default setting adopts 4 neighbor views and the top 8 detected objects in each neighbor view (we also tried other settings, please see Sec~\ref{sec_ablation}).
For simplicity, the objects' positions are roughly represented by their corresponding views' orientations in Eq~\eqref{form_rel_anlge}. This is reasonable as instructions usually mention approximate relative relations of objects and candidates. 

Our model is trained with the widely used two-stage strategy~\cite{tan2019envdrop}. At the first stage, only real training data is used.
At the second stage, we pick the model with the highest SR at the first stage, and keep training it with both real data and synthetic data generated from~\cite{fried2018speaker}.
As for R4R experiment, we only apply the first stage training. We set the projection dimension $\mathbf{D}_k$ in Sec~\ref{sec_predictor} as 512, the glove embedding dimension as 300, and train the agent using RMSprop optimizer~\cite{ruder2016optim} with $1\times 10^{-4}$ learning rate. 
We train the first stage for 80,000 iterations and the second stage will continue training up to 200,000 iterations. All our experiments are conducted on a Nvidia V100 Tensor Core GPU.

\subsection{Comparison with the State-of-The-Art}
We compare NvEM with several SoTA methods under the single-run setting on both R2R and R4R benchmarks. 
Note that VLN mainly focus on the agent's performance on unseen splits, so the performance we report is based on the model which has the highest {SR} on the validation unseen split.

As shown in Table~\ref{table_r2r}, on the R2R benchmark  our NvEM outperforms the baseline EnvDrop~\cite{tan2019envdrop} by a large margin, which obtains $7\%$ absolute improvements in terms of SPL on both Val Unseen and Test splits.
Compared to the state-of-the-art method RelGraph~\cite{hong2020relgraph}, our model obtains $2\%$ absolute improvements on Val Unseen and Test splits. 
Moreover, NvEM even beats some pre-training methods, such as PRESS~\cite{li2019press}, PREVALENT~\cite{hao2020prevelant}, and VLNBert (init. OSCAR)~\cite{hong2020vln-bert}. 
We also note that our method does not surpass VLNBert pre-trained on PREVALENT, which uses in-domain data for pretraining.
As shown later in the ablation study, the success of our model mainly benefits from  the  incorporating of visual contexts, where NvEM effectively improve the textual-visual matching leading to more accurate actions. 

On the R4R benchmark, we observe similar phenomenon with that on R2R. As shown in Table~\ref{table_r4r}, NvEM not only significantly outperforms the baseline EnvDrop~\cite{tan2019envdrop} with $10\%$ absolute improvement, but also sets the new SoTA. 
In particular, NvEM achieves 0.41 CLS, 0.36 nDTW and 0.20 sDTW, which are largely higher than the second best RelGraph~\cite{hong2020relgraph} by 4\%, 4\% and \%2, respectively. 

\subsection{Ablation Study}\label{sec_ablation}
We conduct  ablation experiments over different components of NvEM on R2R dataset. Specifically, we study how the \textit{action}-, \textit{subject}- and \textit{reference}-module contribute to navigation. Then we compare the single-view \textit{subject} module against the neighbor-view one. Lastly, we study how the numbers of neighbor views and objects in the \textit{subject}-and \textit{reference}-module affect the performance. All our ablation models are trained from scratch by two stage training.

\noindent\textbf{The importance of different modules.} Our full model utilizes three modules: the \textit{action}-, \textit{subject}- and \textit{reference}-module, which correspond to  orientations, global views and local objects. 
To study how they affect the performance of navigation, we conduct an ablation experiment by removing corresponding module. The results are shown in Table~\ref{table_phrases}.  
In model \#1, we remove the \textit{action}-module, and it achieves the worst performance compared to others in terms of the main metric SPL. This indicates that orientations are strong guidance for navigation. The same phenomenon is observed in~\cite{hu2019looking,qi2020oaam}. 
In model \#2, we remove  the \textit{subject}-module, and it performs slight better than model\#1 but still far behind the full model. This indicates the global view information is important for VLN.
In model \#3, we remove the \textit{reference}-module, and it performs slight worse than the full model.  This indicates \textit{reference} contains some useful information but not as important as \textit{subject}.
Another reason is that the local \textit{reference} can also be included in global views of the \textit{subject}-module.
\begin{table}[!htbp]
	\caption{Ablation experiment about importance of different modules.}
	\label{table_phrases}
	\vspace{-3mm}
	\centering
	\resizebox{0.48\textwidth}{!}{\begin{tabular}{cccccccc}
			\toprule
			\multicolumn{1}{c}{} & \multicolumn{3}{c}{Modules} & \multicolumn{2}{c}{Val Seen} & \multicolumn{2}{c}{Val Unseen}\\
			\cmidrule(r){2-4} \cmidrule(r){5-6} \cmidrule(r){7-8}
			\multicolumn{1}{c}{model} & \multicolumn{1}{c}{action} & \multicolumn{1}{c}{subject} & \multicolumn{1}{c}{reference} & \multicolumn{1}{c}{SR} & \multicolumn{1}{c}{SPL} & \multicolumn{1}{c}{SR} & \multicolumn{1}{c}{SPL}\\
			\midrule
			1 & & \checkmark & \checkmark & 0.623 & 0.584 & 0.486 & 0.440 \\
			2 & \checkmark & & \checkmark & 0.580 & 0.523 & 0.504 & 0.446 \\
			3 & \checkmark & \checkmark &  & 0.666 & 0.636 & 0.579 & 0.539 \\
			4 & \checkmark & \checkmark & \checkmark & \textbf{0.686} & \textbf{0.645} & \textbf{0.601} & \textbf{0.549} \\
			\bottomrule
	\end{tabular}}
\end{table}

\noindent\textbf{Single-view subject vs neighbor-view subject.} In the \textit{subject}-module, we aggregate neighbor view features based on their spatial affinities to the candidate, which raises the following questions: is neighbor-view more effective than single-view? How about aggregating views in other manners? 
To answer the above questions, we test different types of \textit{subject}-modules, and the results are shown in Table~\ref{table_subject}. Note that we remove the \textit{reference} module to exclude the affect from \textit{reference}.
In table~\ref{table_subject}, \textit{single} uses single-view global features to represent subject,
\textit{lang} uses subject-aware phrases $\mathbf{u}_s$ as query and view features as keys to ground views in Eq~\eqref{form_query}, while \textit{spa} is our default setting which is based on spatial affinity. 
The results show that both model \#2 and \#3 perform better than the single view based model \#1, which indicates the superior of neighbor-view based models. 
Moreover, we observe that language grounded neighbor-view fusion (model \#2) performs worse than the spatial based one (model \#3). This may be caused by the gap between language embeddings and visual embeddings.

\begin{table}[htbp]
	\caption{Different types of \textit{subject}-modules, \textit{single} denotes single-view, \textit{lang} denotes aggregating views using language as query while \textit{spa} denotes using spatial affinity.}
	\label{table_subject}
	\vspace{-3mm}
	\centering
	\resizebox{0.48\textwidth}{!}{\begin{tabular}{cccccccc}
			\toprule
			\multicolumn{1}{c}{} & \multicolumn{3}{c}{Subject} & \multicolumn{2}{c}{Val Seen} & \multicolumn{2}{c}{Val Unseen}\\
			\cmidrule(r){2-4} \cmidrule(r){5-6} \cmidrule(r){7-8}
			\multicolumn{1}{c}{model} & \multicolumn{1}{c}{single} & \multicolumn{1}{c}{lang} & \multicolumn{1}{c}{spa} & \multicolumn{1}{c}{SR} & \multicolumn{1}{c}{SPL} & \multicolumn{1}{c}{SR} & \multicolumn{1}{c}{SPL}\\
			\midrule 
			1 & \checkmark & & & 0.641 & 0.605 & 0.556 & 0.507 \\
			2 & & \checkmark & & 0.637 & 0.608 & 0.564 & 0.522 \\
			3 & & & \checkmark & \textbf{0.666} & \textbf{0.636} & \textbf{0.579} & \textbf{0.539} \\
			\bottomrule
	\end{tabular}}
\end{table}

\begin{figure}[!htbp]
	\centering
	\includegraphics[width=0.45\textwidth]{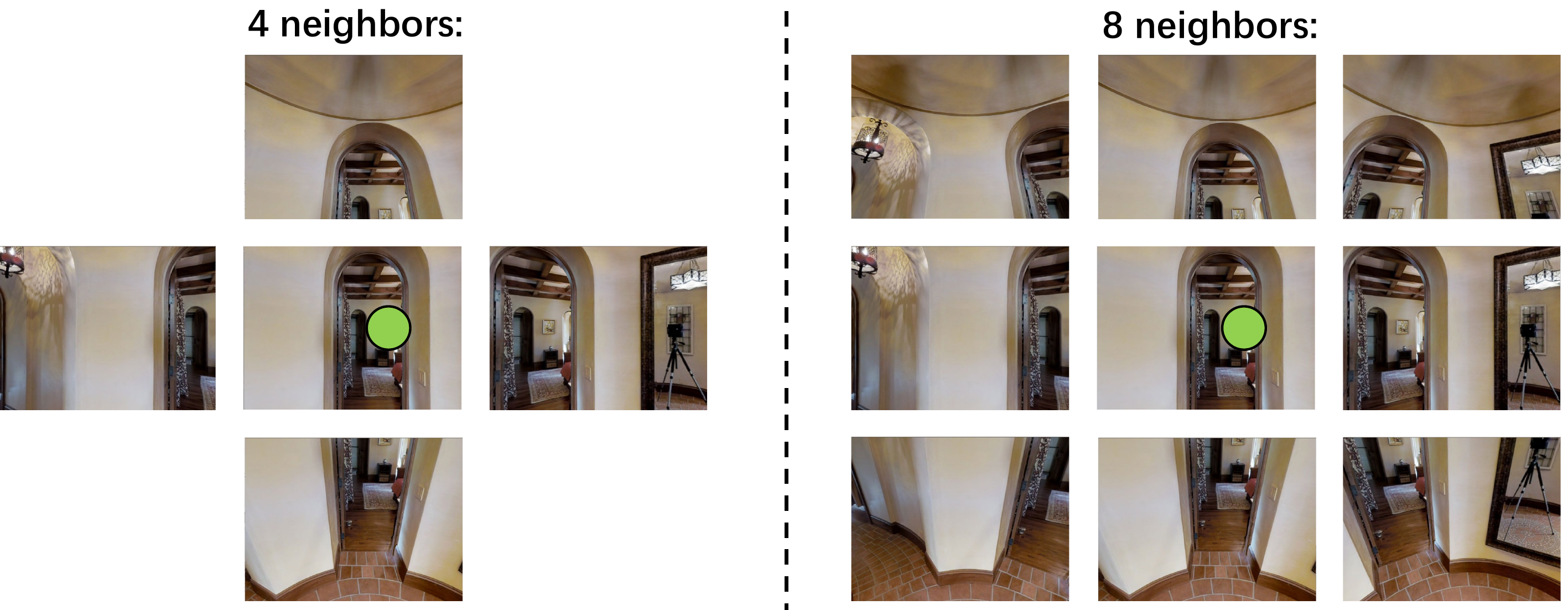}
	\vspace{-3mm}
	\caption{A schematic diagram of 4 neighbors and 8 neighbors.}
	\label{fig_neighbors}
\end{figure}

\noindent\textbf{The analysis of number of neighbors and objects.} In our default setting, the \textit{subject}-module uses 4 neighbor views and the \textit{reference}-module uses 8 objects in each neighbor view. 
However, it is naturally to consider more neighbors (e.g., 8 neighbors) and more objects. In this experiment, we test 4, 8 neighbors, and 4, 8, 12 objects in each view. The results are shown in Table~\ref{table_num}. 
In model \#1 and model \#2, we adjust the number of objects and keep 4 neighbors. They perform worse than our default setting (model \#4). The reason may be  that fewer objects might not contain the objects mentioned by instructions, and more objects could be redundant. 
To study the number of neighbors, we keep the object number as 8. Comparing model \#3 with model \#4, the results show 4 neighbors is the better. The reason   may be that more neighbors contain more redundant information. Not only will they affect the aggregation for subject context, but also increase the difficulty to highlight relevant objects. %
An example is shown in Figure~\ref{fig_neighbors}, intuitively, 8 neighbors have more redundant visual information than 4 neighbors one.

\begin{table}[!htbp]
	\caption{Ablation experiments about the number of neighbor views and objects. \textit{Views} denote the number of neighbors, and \textit{Objects} denote the number of objects in a view.}
	\label{table_num}
	\vspace{-3mm}
	\centering
	\resizebox{0.48\textwidth}{!}{\begin{tabular}{cccccccccc}
			\toprule
			\multicolumn{1}{c}{} & \multicolumn{2}{c}{Views} & \multicolumn{3}{c}{Objects} & \multicolumn{2}{c}{Val Seen} & \multicolumn{2}{c}{Val Unseen}\\
			\cmidrule(r){2-3} \cmidrule(r){4-6} \cmidrule(r){7-8} \cmidrule(r){9-10}
			\multicolumn{1}{c}{model} & 4 & 8 & 4 & 8 & 12 & \multicolumn{1}{c}{SR} & \multicolumn{1}{c}{SPL} & \multicolumn{1}{c}{SR} & \multicolumn{1}{c}{SPL}\\
			\midrule 
			1 & \checkmark & & & & \checkmark & 0.655 & 0.617 & 0.564 & 0.517 \\
			2 & \checkmark & & \checkmark & & & 0.667 & 0.638 & 0.572 & 0.532 \\
			\midrule
			3 & & \checkmark & & \checkmark & & 0.644 & 0.615 & 0.564 & 0.523 \\
			4 & \checkmark & & & \checkmark & & \textbf{0.686} & \textbf{0.645} &
			\textbf{0.601} & \textbf{0.549} \\
			\bottomrule
	\end{tabular}}
\end{table}
\vspace{-5mm}
\begin{figure*}[!t]
	\centering
	\includegraphics[width=0.9\textwidth]{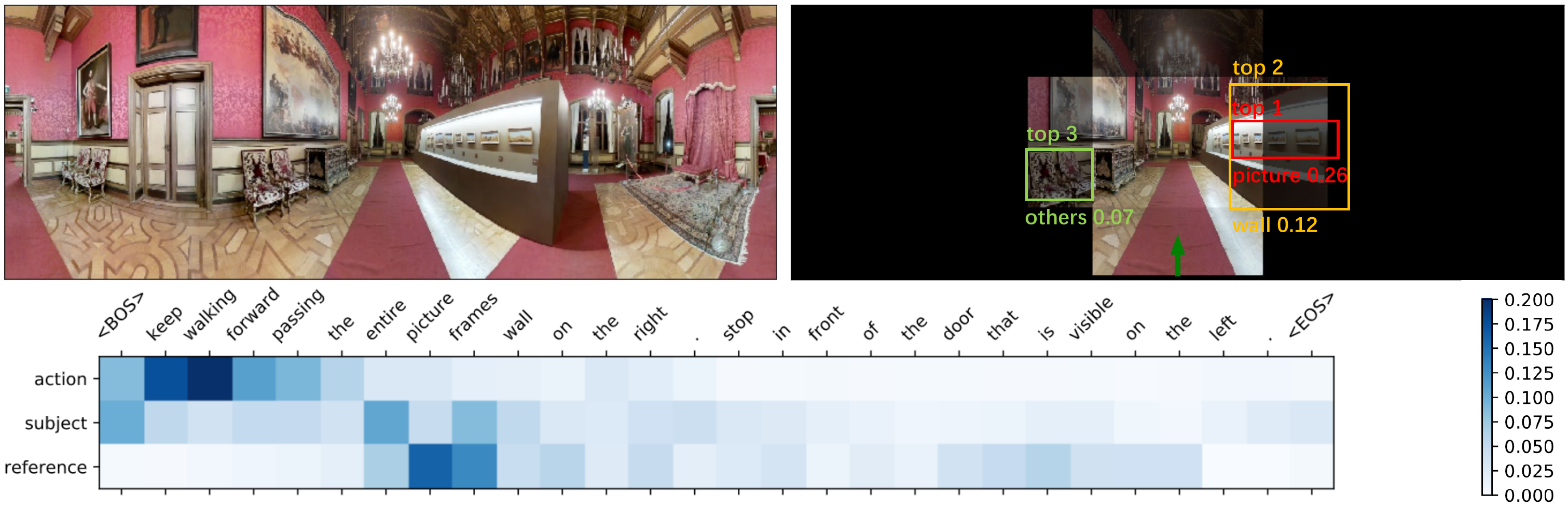}
	\includegraphics[width=0.9\textwidth]{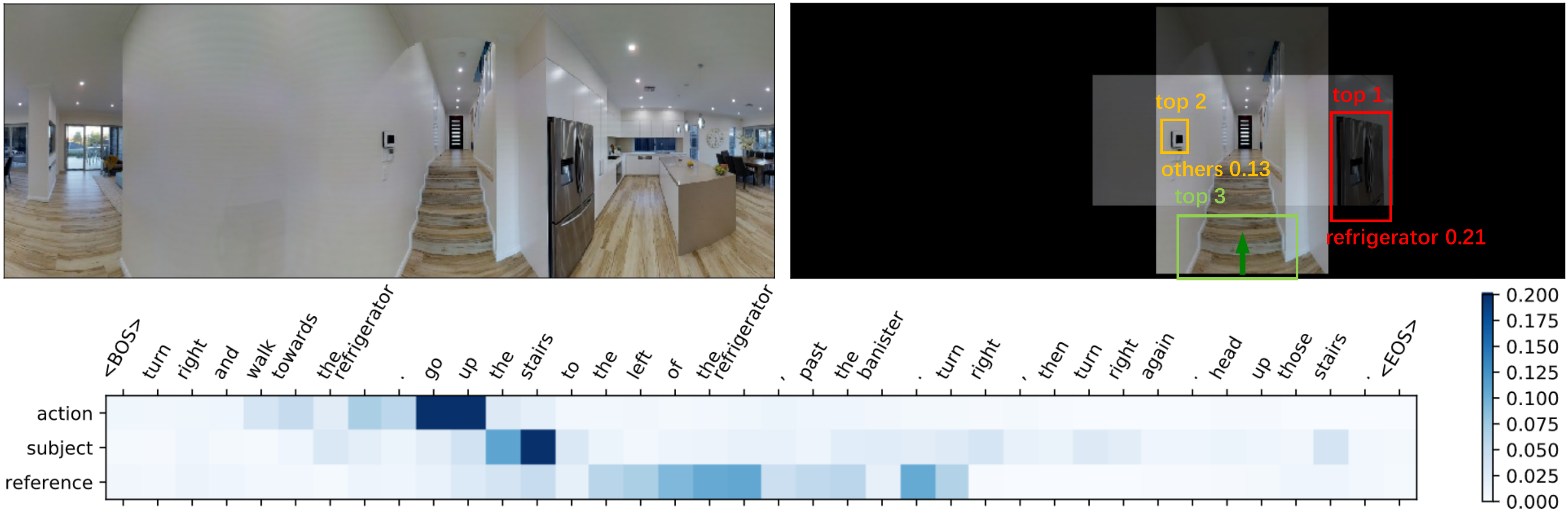}
	\vspace{-3mm}
	\caption{Visualizations of two success cases. The mask on the panorama denotes neighbor-view spatial affinities of the predicted candidate in \textit{subject} module (note that neighbor views overlap with the center view). The boxes are objects with top three scores in \textit{reference} module. Attentions on three types of phrases are also showed.}
	\label{fig_success}
\end{figure*}
\begin{figure*}[!t]
	\centering
	\includegraphics[width=0.9\textwidth]{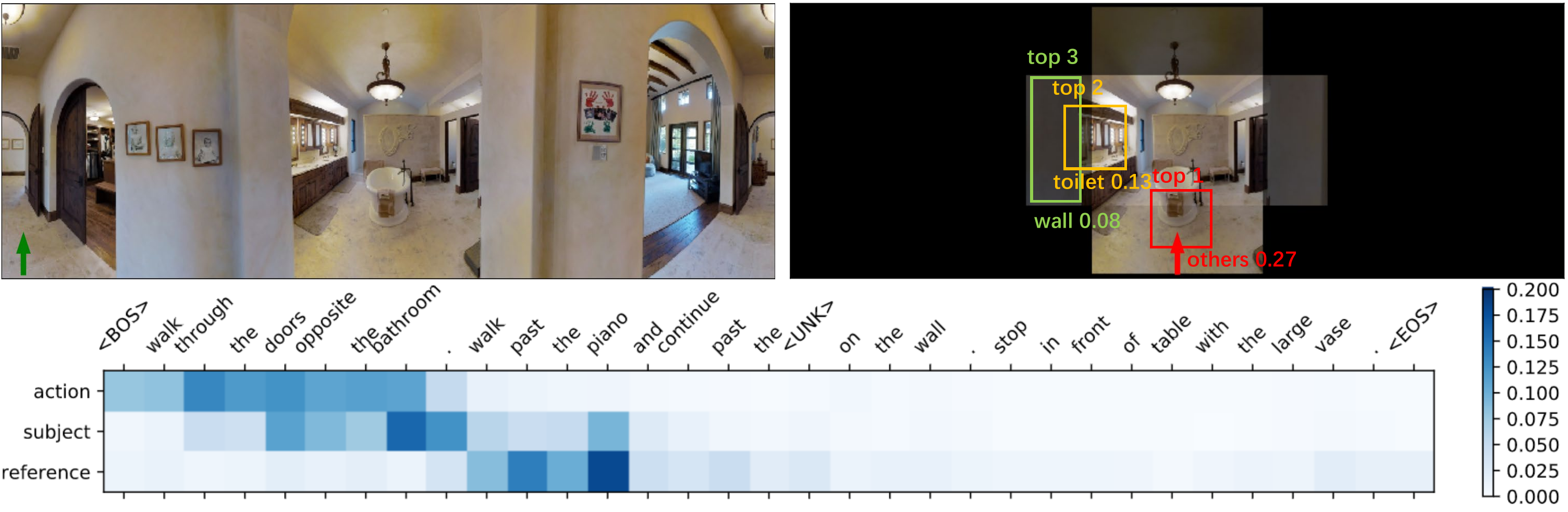}
	\vspace{-3mm}
	\caption{Visualization of a failure case. Green arrow denotes ground-truth action, while red arrow denotes predicted action. The mask on the panorama denotes neighbor-view spatial affinities of the predicted candidate in \textit{subject} module (note that neighbor views overlap with the center view). The boxes are objects with top three scores in \textit{reference} module. Attentions on three types of phrases are also showed.}
	\label{fig_failure}
\end{figure*}

\vspace{3mm}
\subsection{Qualitative Visualizations}
Here we present some success and failure cases.
Figure~\ref{fig_success} shows two success cases. Taking the bottom one as an example, each module attends on correct phrases, such as \textit{``go up''}, \textit{``stairs''} and \textit{``to the left of the refrigerator''} corresponding to \textit{action}, \textit{subject} and \textit{reference} modules respectively. 
More visual information of the subject \textit{``stairs''} from neighbor views are incorporated by our \textit{subject}-module. 
In addition, taking advantage of the \textit{reference}-module, our model could perceive the mentioned \textit{``refrigerator''} in neighbor views as reference, and suppress unmentioned ones.
We note that the attention scores of objects are not that high, which may mainly caused by the fact that we consider totally 32 objects  for each candidate (4 neighbors and 8 objects in each neighbor). 

Figure~\ref{fig_failure} visualizes a failure case. The model mainly focus on the \textit{``bathroom''} in the \textit{subject}-module, which leads to a wrong direction. This indicates that the attentive phrases of NvEM are not always correct, and thus it still could be improved by studying how to extract more accurate phrases.

\section{Conclusion}
In this paper, we present a novel multi-module Neighbor-view Enhanced Model to improve textual-visual matching via adaptively incorporating   visual information from neighbor views. Our subject module   aggregates neighbor views at a global level based on spatial-affinity, and our reference module aggregates neighbor objects at a local level guided by referring phrases.
Extensive experiments demonstrate that NvEM effectively improves the agent's performance in unseen environments and our method sets the new state-of-the-art. 

Considering the similarity between R2R task and other VLN tasks, such as dialogue navigation~\cite{thomas2019dialogue,khanh2019help}, remote object detection~\cite{qi2020reverie} and navigation in continuous space~\cite{jacob2020vlnce}, we believe the neighbor-view enhancement idea could also benefit agents in other embodied AI tasks. We leave them as future works to explore.

\section{Acknowledgments}
This work was jointly supported by National Key Research and Development Program of China Grant No. 2018AAA0100400, National Natural Science Foundation of China (61525306, 61633021, 61721004, 61806194, U1803261, and 61976132), Beijing Nova Program (Z201100006820079), Shandong Provincial Key Research and Development Program (2019JZZY010119), Key Research Program of Frontier Sciences CAS Grant No.ZDBS-LY-JSC032, and CAS-AIR.

\bibliographystyle{ACM-Reference-Format}
\bibliography{an}

\end{document}